\documentclass[10pt]{book}

\usepackage[dvips]{graphicx}
\usepackage[english]{babel}
\usepackage[latin1]{inputenc}
\usepackage{amsmath} 
\usepackage{amssymb}

\def\rien{\rule{0pt}{0pt}}
\def\Rset{\mathrm{I\!R}}
\def\genFrame#1{G^{#1}}

\begin{document}

\chapter{Definition of evidence fusion rules based on referee functions}
Fr\'ed\'eric Dambreville\vspace{3pt}\\
D\'el\'egation G\'en\'erale pour l'Armement,\\
DGA/CEP/EORD/FAS,\\
16 bis, Avenue Prieur de la C\^ote d'Or\\
Arcueil, F 94114, France\\[3pt]
Form mail: http://email.fredericdambreville.com
\\\\
{\bf Abstract.}
This chapter defines a new concept and framework for constructing fusion rules for evidences.
This framework is based on a \emph{referee function}, which does a decisional arbitrament conditionally to basic decisions provided by the several sources of information.
A simple sampling method is derived from this framework.
The purpose of this sampling approach is to avoid the combinatorics which are inherent to the definition of fusion rules of evidences.
This definition of the fusion rule by the means of a sampling process makes possible the construction of several rules on the basis of an algorithmic implementation of the referee function, instead of a mathematical formulation.
Incidentally, it is a versatile and intuitive way for defining rules.
The framework is implemented for various well known evidence rules.
On the basis of this framework, new rules for combining evidences are proposed, which takes into account a consensual evaluation of the sources of information. 
\\\\
{\bf Keywords: Evidence, Referee Function, Sampling, Dempster-Shafer rule, PCR6.}
\section*{Notations}
\begin{itemize}
\item $I[b]$, function of Boolean $b$, is defined by $I[\mathbf{true}]=1$ and $I[\mathbf{false}]=0$\,.
Typically, $I[x=y]$ has value $1$ when $x=y$, and $0$  when $x\ne y$,
\item Let be given a frame of discernment $\Theta$.
Then, the structure $\genFrame\Theta$ denotes any \emph{lattice} constructed from $\Theta$.
In particular, $\genFrame\Theta$ may be a distributive lattice like the hyper-power set $D^\Theta$;
or $\genFrame\Theta$ may be a Boolean algebra like the power set $2^\Theta$, the \emph{superpower set} $S^\Theta$, or the free Boolean algebra generated by $\Theta$,
\item $x_{1:n}$ is an abbreviation for the sequence $x_1,\cdots,x_n$\,,
\item $\max\{x_1,\cdots,x_n\}$, or $\max\{x_{1:n}\}$\,, is the maximal value of the sequence $x_{1:n}$\,.
Similar notations are used for $\min$\,,
\item  $\max_{x\in X}\{f(x)\}$, or $\max\{f(x)\;/\;x\in X\}$, is the maximal value of $f(x)$ when $x\in X$\,.
Similar notations are used for $\min$\,
\end{itemize}
\section{Introduction}
Evidence theory~\cite{Dempster1967,Shafer1976} has often been promoted as an alternative approach for fusing information, when the hypotheses for a Bayesian approach cannot be precisely stated. 
While many academic studies have been accomplished, most industrial applications of data fusion still remain based on a probabilistic modeling and fusion of the information. 
This great success of the Bayesian approach is explained by at least three reasons:
\begin{itemize}
\item The underlying logic of the Bayesian inference~\cite{BayesianLogic} seems intuitive and obvious at a first glance.
It is known however~\cite{lewis} that the logic behinds the Bayesian inference is much more complex,
\item The Bayesian rule is entirely compatible with the preeminent theory of Probability and takes advantage of all the probabilistic background, 
\item Probabilistic computations are tractable, even for reasonably complex problems. 
\end{itemize}
Then, even if evidences allow a more general and subtle manipulation of the information for some case of use, the Bayesian approach still remains the method of choice for most applications.
This chapter intends to address the three afore mentioned points, by providing a random set interpretation of the fusion rules.
This interpretation is based on a \emph{referee function}, which does a decisional arbitrament conditionally to basic decisions provided by the several sources of information.
This referee function will imply a sampling approach for the definition of the rules.
Sampling approach is instrumental for the combinatorics avoidance~\cite{Papoulis1984}.
\\[5pt]
In the recent literature, there has been a large amount of work devoted to the definition of new fusion rules~\cite{Denoeux2006} to~\cite{Yager1987}\,.
The choice for a rule is often dependent of the applications and there is not a systematic approach for this task. 
The definition of the fusion rule by the means of a sampling process makes possible the construction of several rules on the basis of an algorithmic implementation of the referee function, instead of a mathematical formulation.
Incidentally, it is a versatile and intuitive way for defining rules.
Subsequently, our approach is illustrated by implementing two well known evidence rules.
On the basis of this framework, new rules for combining evidences are also proposed.
Typically, these new rules takes into account a consensual evaluation of the sources, by invalidating irrelevant sources of information on the basis of a majority decision.
\\[5pt]
Section~\ref{DSmB3:2:F2K9:section:1.5} summarize some classical results in the domain of evidence.
Section~\ref{DSmB3:2:F2K9:section:2} introduces the notion of referee function and its application to the definition of fusion rules. A sampling method is obtained as a corollary.
Section~\ref{DSmB3:2:F2K9:section:3} establishes the referee functions for two known rules.
Section~\ref{DSmB3:2:F2K9:section:4} defines new fusion rules.
Section~\ref{DSmB3:2:f2k9:F2K9:sect:Numexample} makes some numerical comparisons.
Section~\ref{DSmB3:2:F2K9:sect:conclude} concludes.
%
\section{Belief fusion}
\label{DSmB3:2:F2K9:section:1.5}
This section introduces the notion of belief function and some classical rules of fusion.
\subsection{Lattices}
Lattices are algebraic structures which are useful for encoding logical information.
In particular, lattices are generalizations of structures like Boolean algebra, sets, hyper-power sets or \emph{concept lattice}~\cite{latticeTheory,FormalConceptAnalysis}.
Sets and hyper-power sets are widely used as a framework for defining and manipulating belief functions.
Concept lattices are frameworks used in the ontologic domain of \emph{formal concept analysis}.

\paragraph{Definition 1.}
A (finite) \emph{lattice} $L$ is a partially ordered (finite) set, \emph{i.e.} (finite) poset, which satisfies the following properties:
\begin{itemize}
\item For any two elements $X,Y\in L$, there is a greatest lower bound $X\wedge Y$ of the set $\{X, Y\}$\,,
\item For any two elements $X,Y\in L$, there is a least upper bound $X\vee Y$ of the set $\{X, Y\}$\,.
\end{itemize}
The notations $X\cap Y$ (respectively $X\cup Y$) are also used instead of $X\wedge Y$ (respectively $X\vee Y$).
\paragraph{Example.}
Concept lattices~\cite{FormalConceptAnalysis} (which are not defined here) are lattices.
Concept lattices are widely used for deriving ontologies.
\paragraph{Definition 2.}
A \emph{bounded lattice} $L$ is a lattice which have a least element $\bot$ and a greatest element $\top$.
\\[5pt]
The notations $\emptyset$ or $0$ (respectively $\Omega$ or $1$) are also used instead of $\bot$ (respectively $\top$).
\paragraph{Proposition.}
A finite lattice is a bounded lattice.
\\[5pt]
\emph{Proof is achieved by setting:}
$
\bot=\bigwedge_{X\in L}X
\quad\mbox{and}\quad
\top=\bigvee_{X\in L}X
\;.
$
\paragraph{Definition 3.}
A \emph{distributive lattice} $L$ is a lattice such that $\wedge$ and $\vee$ are mutually distributive:
$$
X\wedge (Y\vee Z)=(X\wedge Y)\vee (X\wedge Z)
\quad\mbox{and}\quad
X\vee (Y\wedge Z)=(X\vee Y)\wedge (X\vee Z)\;,
$$
for any $X,Y,Z\in L\,.$
\paragraph{Example.}
Being given a frame of discernment $\Theta$, the hyper-power set $D^\Theta$ is typically the free distributive lattice generated by $\Theta$.
\paragraph{Definition 4.}
A \emph{complemented lattice} is a bounded lattice $L$, such that each element $X\in L$ has a \emph{complement}, \emph{i.e.} an element $Y\in L$ verifying:
$$
X\vee Y=\top
\quad\mbox{and}\quad
X\wedge Y=\bot\;.
$$
The complement of $X$ is often denoted $\neg X$ or $X^c$\,.
\\[5pt]
Actually, the complementation is defined by introducing constraints on the lattice.
As a consequence, \emph{it is possible to have partial complementation on the lattice.}
\paragraph{Definition 5.}
A Boolean algebra is a complemented distributive lattice.
\paragraph{Examples.}
Being given a frame of discernment $\Theta$, the power set $2^\Theta$, the \emph{superpower set} $S^\Theta$, or the free Boolean algebra generated by $\Theta$, are Boolean algebra.
\paragraph{Conclusion.}
Bounded lattices (especially, finite lattices) are versatile structures, which are able to address many kind of informational frameworks.
Typically, bounded lattices generalize power set and hyper-power set.
But since complementation is defined by introducing constraints on the lattice, it is also possible to derive lattice with partial complementation (\emph{i.e.} with a subset of the complementation constraints) which are intermediate structure between hyper-power set and power set.
Since bounded lattices are such generalization, this chapter will define evidence fusion rules within this framework.
By using a lattice framework, we are also linking our work to the domain of \emph{formal concept analysis}.
\\[5pt]
In the domain of evidence theories, basic concepts are generally modeled by means of a set of proposition, $\Theta$, called \emph{frame of discernment}.
The structure $\genFrame\Theta$ is any finite lattice constructed from $\Theta$\,.
In particular, $\genFrame\Theta$ may be $D^\Theta$, $2^\Theta$ or $S^\Theta$.
Notice that $\genFrame\Theta$ is not a lost of generality, and is able to address any finite lattice.
From now on, all notions and results are defined within $\genFrame\Theta$, which make them applicable also to any finite lattice.
\subsection{Belief functions}
Belief functions are measures of the uncertainty that could be defined on the propositions of a bounded lattice.
In the framework of evidence theory, beliefs are defined so as to manage not only the uncertainty, but also the imprecision encoded within the lattice structure.
\paragraph{Basic belief assignment.}
A basic belief assignment (bba) on $\genFrame\Theta$ is a mapping $m:\genFrame\Theta\rightarrow \Rset$ such that:
\begin{itemize}
\item $m\ge0$\;,
\item $m(\emptyset)=0$\;,
\item $\displaystyle\sum_{X\in\genFrame\Theta}m(X)=1$\;.
\end{itemize}
The bba contains an elementary knowledge (in the form of a basic belief) about the whole propositions of $\genFrame\Theta$.
The bba, however, does not provide directly the knowledge about an individual proposition.
This individual knowledge is imprecise and bounded by the \emph{belief} and the \emph{plausibility}.
\paragraph{Belief.}
The belief $\mathrm{bel}$ is constructed from the bba $m$ as follows:
$$
bel(X)=\sum_{\substack{Y\in\genFrame\Theta\\Y\subset X}}m(X)\;.
$$
The belief is a pessimistic interpretation of the bba.
\paragraph{Plausibility.}
The plausibility $\mathrm{pl}$ is constructed from the bba $m$ as follows:
$$
pl(X)=\sum_{\substack{Y\in\genFrame\Theta\\Y\cap X\ne\emptyset}}m(X)\;.
$$
The plausibility is an optimistic interpretation of the bba.
\\[8pt]
While fusing beliefs, the essential computations are done by means of the bbas.
Our contribution is focused on the bba fusion; belief and plausibility will not be manipulated in this chapter.
Notice howover that some properties of the belief functions may change, depending on the structure of the lattice being used. 
\subsection{Fusion rules}
Let be given $s$ sources of information characterized by their bbas $m_{1:s}$. 
How could we fuse these information into a single fused bba?
There is a variety of rules for fusing bba.
This section covers different classical rules.
\paragraph{Dempster-Shafer.}
The fused bba $m_{\text{\tiny{DST}}}$ obtained from $m_{1:s}$ by means of \emph{Dempster-Shafer} fusion rule~\cite{Dempster1967,Shafer1976} is defined by:
\begin{equation}\label{DSmB3:2:eqfond:1}\begin{array}{@{}l@{}}
m_{\text{\tiny{DST}}}(\emptyset)=0
%
\mbox{ \ and \ }
m_{\text{\tiny{DST}}}(X)=\frac{\displaystyle
m_{\wedge}(X)
}{\displaystyle
1-m_{\wedge}(\emptyset)
}\mbox{ \ for any }X\in\genFrame\Theta\setminus\{\emptyset\}\;.
\end{array}\end{equation}
where $m_{\wedge}(\cdot )$ corresponds to the conjunctive consensus:
\begin{equation}\label{DSmB3:2:eqfond:2}m_{\wedge}(X)\triangleq \sum_{\substack{
Y_1\cap\cdots\cap Y_s=X
 \\ 
 Y_1,\cdots,Y_s\in \genFrame\Theta
}}\prod_{i=1}^s m_{i}(Y_i)\;,\end{equation}
 for any $X\in\genFrame\Theta\setminus\{\emptyset\}\;.$
\\[5pt]
 The rejection rate $z=m_{\wedge}(\emptyset)$ is a measure of the conflict between the sources.
 Notice that the conflict is essentially a conjunctive notion, here.
\\[5pt]
Historically, this is the first  rule for fusing evidences.
Essentially, this rule provides a cross fusion of the information: it is based on a conjunctive kernel.
However, the conjunctive nature of this rule is altered by the necessary normalization implied by the conflict measurement $m_{\wedge}(\emptyset)$.
\subsection{Disjunctive rule}
The fused bba $m_{\vee}$ obtained from $m_{1:s}$ by means of a \emph{disjunctive} fusion is defined by:
\begin{equation}\label{DSmB3:2:eqfond:2:b}
m_{\vee}(X)=\sum_{\substack{
Y_1\cup\cdots\cup Y_s=X
 \\ 
 Y_1,\cdots,Y_s\in \genFrame\Theta
}}\ \prod_{i=1}^s m_{i}(Y_i)\;,\end{equation}
 for any $X\in\genFrame\Theta\;.$
\\[5pt]
The disjunctive rule alone is not very useful, but it is interesting when fusing highly conflicting information.
When at least one sensor provides the good answer, the disjunctive rule will maintain a minimal knowledge. 
Typically, this rule may be combined with the conjunctive consensus, in an adaptive way~\cite{Florea2006,Dubois}.

\paragraph{Rule of Dubois $\&$ Prade.}
Dempster-Shafer fusion rule will have some unsatisfactory behavior, when the conflict level is becoming high.
If it is assumed that at least one sensor provides the good answer, then it is wiser to replace the possible conflict by a disjunctive repartition of the belief product.
This idea is implemented by the rule of Dubois and Prade~\cite{Dubois}.
\\[5pt]
The fused bba $m_{\text{\tiny{D\&P}}}$ obtained from $m_{1:s}$ by means of this fusion rule is defined by:
\begin{equation}\label{DSmB3:2:eqfond:3}\begin{array}{@{}l@{}}
m_{\text{\tiny{D\&P}}}(\emptyset)=0
\;,
\\\displaystyle
m_{\text{\tiny{D\&P}}}(X)=
\sum_{\substack{
Y_1\cap\cdots\cap Y_s=X
 \\ 
 Y_1,\cdots,Y_s\in \genFrame\Theta
}}\ \prod_{i=1}^s m_{i}(Y_i)
+
\sum_{\substack{
Y_1\cap\cdots\cap Y_s=\emptyset
 \\ 
Y_1\cup\cdots\cup Y_s=X
 \\ 
 Y_1,\cdots,Y_s\in \genFrame\Theta
}}\ \prod_{i=1}^s m_{i}(Y_i)
\;,
\end{array}\end{equation}
for any $X\in\genFrame\Theta\setminus\{\emptyset\}\;.$
\paragraph{Averaging rule.}
Averaging, although quite simple, may provide good results on some applications.
\\[5pt]
Let be given the averaging parameters $\alpha_{1:s}\ge0$ such that $\sum_{i=1}^s\alpha_i=1$\,.
The averaged bba $m_{\mu}[\alpha]$ is obtained from $m_{1:s}$ by:
$$
m_{\mu}[\alpha]=\sum_{i=1}^s \alpha_i m_i\;.
$$
\paragraph{PCR6.}
The proportional conflic redistribution rules (PCR$n$) have been introduced By Smarandache and Dezert \cite{Smarandache2006}.
The rule PCR6 has been proposed by Martin and Osswald in~\cite{Martin2006}\,.
PCR rules will typically replace the possible conflict by an adaptive averaging of the belief product.
\\[5pt]
The fused bba $m_{\text{\tiny{PCR6}}}$ obtained from $m_{1:s}$ by means of PCR6 is defined by:
$$
m_{\text{\tiny{PCR6}}}(\emptyset)=0\;,
$$
and, for any $X\in\genFrame\Theta\setminus\{\emptyset\}\;,$ by:
\begin{equation}\label{DSmB3:2:eqfond:4}\begin{array}{@{}l@{}}\displaystyle
m_{\text{\tiny{PCR6}}}(X)=m_{\wedge}(X)
\vspace{8pt}\\\displaystyle
\rien\qquad+
\sum_{i=1}^s m_i(X)^2\sum_{\substack{
\bigcap_{k=1}^{s-1} Y_{\sigma_i(k)}\cap X = \emptyset
 \\
 Y_{\sigma_i(1)},\cdots,Y_{\sigma_i(s-1)}\in \genFrame\Theta
 }}
 \left(\frac{\displaystyle
\prod_{j=1}^{s-1}
m_{\sigma_i(j)}(Y_{\sigma_i(j)}) 
 }{\displaystyle
m_i(X)+
\sum_{j=1}^{s-1}
m_{\sigma_i(j)}(Y_{\sigma_i(j)}) 
 }\right)\;,
\end{array}\end{equation}
\noindent
where the function $\sigma_i$ counts from $1$ to $s$ avoiding $i$\,:
$$
    \sigma_i(j)=j\times I[j<i] + (j+1)\times I[j\ge i]\;.
$$
\emph{N.B. If the denominator in~(\ref{DSmB3:2:eqfond:4}) is zero, then the fraction is discarded.}
\paragraph{Prospective.}
The previous rules are just examples amongst many possible rules.
Most of the rules are characterized by their approaches for handling the conflict.
Actually, there is not a definitive criterion for the choice of a particular rule.
There is not a clearly intuitive framework for the comparison of the rules as well.
This chapter addresses this diversity by proposing a constructive interpretation of fusion rules by means of the notion of referee functions. 
%
\section{Referee function and fusion rules}
\label{DSmB3:2:F2K9:section:2}
\subsection{Referee function}
\paragraph{Definition.}
A referee function on $\genFrame\Theta$ for $s$ sources of information and with context $\gamma$ is a mapping $X,Y_{1:s}\mapsto F(X|Y_{1:s};\gamma)$ defined on propositions $X,Y_{1:s}\in\genFrame\Theta$\,, which satisfies:
\begin{itemize}
\item $\displaystyle
F(X|Y_{1:s};\gamma)\ge0\;,$
\item $\displaystyle
\sum_{X\in\genFrame\Theta}F(X|Y_{1:s};\gamma)=1\;,$
\end{itemize}
for any $X,Y_{1:s}\in\genFrame\Theta\,.$
\\[5pt]
A referee function for $s$ sources of information is also called a $s$-ary referee function.
The quantity $F(X|Y_{1:s};\gamma)$ is called a \emph{conditional arbitrament} between $Y_{1:s}$ in favor of $X$.
Notice that $X$ is not necessary one of the propositions $Y_{1:s}$\,; typically, it could be a combination of them.
The case $X=\emptyset$ is called the \emph{rejection case}.
\paragraph{Fusion rule.}
Let be given $s$ basic belief assignments (bba) $m_{1:s}$ and a $s$-ary referee function $F$ with context $m_{1:s}$\,.
Then, the fused bba $m_1\oplus\cdots\oplus m_s[F]$ based on the referee $F$ is constructed as follows: 
\begin{equation}\label{DSmB3:2:F2K9:eq:def}
m_1\oplus\cdots\oplus m_s[F](X)=I[X\ne\emptyset]\frac{\displaystyle
\sum_{Y_{1:s}\in\genFrame\Theta}
F(X|Y_{1:s};m_{1:s})
\prod_{i=1}^s
m_i(Y_i)
}{\displaystyle
1-\sum_{Y_{1:s}\in\genFrame\Theta}
F(\emptyset|Y_{1:s};m_{1:s})
\prod_{i=1}^s
m_i(Y_i)
}\;,
\end{equation}
for any $X\in\genFrame\Theta$\,.
\\[5pt]
From now on, the notation $\oplus[m_{1:s}|F]=m_1\oplus\cdots\oplus m_s[F]$ is used.
\\[5pt]
The value $z=\sum_{Y_{1:s}\in\genFrame\Theta}
F(\emptyset|Y_{1:s};m_{1:s})
\prod_{i=1}^s m_i(Y_i)$ is called the \emph{rejection rate}.
Notice that the rejection rate is derived from the rejection generated by $F(\emptyset|Y_{1:s};m_{1:s})$\,
As a consequence, the rejection is not exclusively a conjunctive notion in this approach.
An example of non conjunctive rejection is proposed in section~\ref{DSmB3:2:F2K9:section:4}.
\paragraph{Examples.}
Refer to section~\ref{DSmB3:2:F2K9:section:3} and~\ref{DSmB3:2:F2K9:section:4}.
\subsection{Properties}
\paragraph{Bba status.}
The function $\oplus[m_{1:s}|F]$ defined on $\genFrame\Theta$ is actually a basic belief assignment.
\\[5pt]
\emph{Proof.}
It is obvious that $\oplus[m_{1:s}|F]\ge0$\,.
\\
Since $I[\emptyset\ne\emptyset]=0$\,, it is derived $\oplus[m_{1:s}|F](\emptyset)=0$\,.
\\
From $\sum_{X\in\genFrame\Theta}F(X|Y_{1:s};m_{1:s})=1
$\,, it is derived:
$$\begin{array}{@{}l@{}}\displaystyle
\sum_{X\in\genFrame\Theta}\sum_{Y_{1:s}\in\genFrame\Theta}
F(X|Y_{1:s};m_{1:s})
\prod_{i=1}^s
m_i(Y_i)
\\[3pt]\displaystyle
\rien\qquad\qquad
=
\sum_{Y_{1:s}\in\genFrame\Theta}
\left(\prod_{i=1}^s
m_i(Y_i)
\right)
\sum_{X\in\genFrame\Theta}F(X|Y_{1:s};m_{1:s})
\\[3pt]\displaystyle
\rien\qquad\qquad
=
\sum_{Y_{1:s}\in\genFrame\Theta}
\ \prod_{i=1}^s
m_i(Y_i)
\\[3pt]\displaystyle
\rien\qquad\qquad
=
\prod_{i=1}^s
\ \sum_{Y_i\in\genFrame\Theta}
m_i(Y_i)
=1\;.
\end{array}$$
As a consequence:
$$\begin{array}{@{}l@{}}\displaystyle
\sum_{X\in\genFrame\Theta}I[X\ne\emptyset]\sum_{Y_{1:s}\in\genFrame\Theta}
F(X|Y_{1:s};m_{1:s})
\prod_{i=1}^s
m_i(Y_i)
\\[3pt]\displaystyle
\rien\qquad\qquad\qquad\qquad\qquad\qquad
+
\sum_{Y_{1:s}\in\genFrame\Theta}
F(\emptyset|Y_{1:s};m_{1:s})
\prod_{i=1}^s
m_i(Y_i)
=
1\;.
\end{array}$$
$\Box\Box\Box$
\paragraph{Referee function without rejection.}
Let be given $s$ basic belief assignments (bba) $m_{1:s}$ and a $s$-ary referee function $F$ with context $m_{1:s}$\,.
Assume that $F$ does not imply rejection, that is:
$$
F(\emptyset|Y_{1:s};m_{1:s})=0\quad\mbox{for any }Y_{1:s}\in\genFrame\Theta\setminus\{\emptyset\}\;.
$$
Then, the fused bba $\oplus[m_{1:s}|F]$ based on the referee $F$ has the simplified definition: 
\begin{equation}\label{DSmB3:2:F2K9:eq:def:bis}
\oplus[m_{1:s}|F](X)=\sum_{Y_{1:s}\in\genFrame\Theta}
F(X|Y_{1:s};m_{1:s})
\prod_{i=1}^s
m_i(Y_i)
\;,
\end{equation}
for any $X\in\genFrame\Theta$\,.
\\[5pt]
\emph{Proof.}
It is a consequence of $\displaystyle \sum_{Y_{1:s}\in\genFrame\Theta}
F(\emptyset|Y_{1:s};m_{1:s})
\prod_{i=1}^s
m_i(Y_i)=0\;.$
\paragraph{Separability.}
Let be given $s$ basic belief assignments (bba) $m_{1:s}$ and a $s$-ary referee function $F$ with context $m_{1:s}$\,.
Assume that there is $r$ and $t$ such that $r+t=s$, and two sequences $u_{1:r}$ and $v_{1:t}$ which constitute a partition of $[\![1,s]\!]$\,, that is:
$$
\{u_{1:r}\}\cup\{v_{1:t}\}=[\![1,s]\!]
\quad\mbox{and}\quad
\{u_{1:r}\}\cap\{v_{1:t}\}=\emptyset\;.
$$
Assume also that there are
a $r$-ary referee function $G$ with context $m_{u_{1:r}}$\;,
a $t$-ary referee function $H$ with context $m_{v_{1:t}}$
and a parameter $\theta\in[0,1]$
such that:
$$
F(X|Y_{1:s};m_{1:s})=
\theta G(X|Y_{u_{1:r}};m_{u_{1:r}})
 + 
(1-\theta) H(X|Y_{v_{1:t}};m_{v_{1:t}})\;.
$$
Then $F$ is said to be \emph{separable} into the \emph{two} sub-referee functions $G$ and $H$.
Moreover, the fused bba is simplified as follows:
\begin{equation}
\label{DSmB3:2:F2K9:eq:separable}
\oplus[m_{1:s}|F]=\theta \oplus[m_{u_{1:r}}|G]+(1-\theta) \oplus[m_{v_{1:t}}|H]
\end{equation}
Notice that the fusion is easier for small arity.
As a consequence, separability provides possible simplifications to the fusion.
\\[5pt]
\emph{Proof.}
It is derived:
$$\begin{array}{@{}l@{}}\displaystyle
\sum_{X\in\genFrame\Theta}\sum_{Y_{1:s}\in\genFrame\Theta}
F(X|Y_{1:s};m_{1:s})
\prod_{i=1}^s
m_i(Y_i)
\\[3pt]\displaystyle
\rien\qquad\qquad
=
\theta\sum_{X\in\genFrame\Theta}\sum_{Y_{1:s}\in\genFrame\Theta}
G(X|Y_{u_{1:r}};m_{u_{1:r}})
\prod_{i=1}^s
m_i(Y_i)
\\[3pt]\displaystyle
\rien\qquad\qquad\quad
+
(1-\theta)\sum_{X\in\genFrame\Theta}\sum_{Y_{1:s}\in\genFrame\Theta}
H(X|Y_{v_{1:t}};m_{v_{1:t}})
\prod_{i=1}^s
m_i(Y_i)
\;.
\end{array}$$
Now:
$$\begin{array}{@{}l@{}}\displaystyle
\sum_{X\in\genFrame\Theta}\sum_{Y_{1:s}\in\genFrame\Theta}
G(X|Y_{u_{1:r}};m_{u_{1:r}})
\prod_{i=1}^s
m_i(Y_i)
\\[5pt]\displaystyle
\rien\qquad
=
\left(\sum_{X\in\genFrame\Theta}\sum_{Y_{u_{1:r}}\in\genFrame\Theta}
G(X|Y_{u_{1:r}};m_{u_{1:r}})
\prod_{i=1}^r
m_{u_i}(Y_{u_i})\right)
\prod_{i=1}^t
\sum_{Y_{v_i}\in\genFrame\Theta}
m_{v_i}(Y_{v_i})
\\[5pt]\displaystyle
\rien\qquad
=
\oplus[m_{u_{1:r}}|G]\prod_{i=1}^t 1=\oplus[m_{u_{1:r}}|G]
\;.
\end{array}$$
It is derived similarly:
$$
\sum_{X\in\genFrame\Theta}\sum_{Y_{1:s}\in\genFrame\Theta}
H(X|Y_{v_{1:t}};m_{v_{1:t}})
\prod_{i=1}^s
m_i(Y_i)
=
\oplus[m_{v_{1:t}}|H]
\;.
$$
$\Box\Box\Box$
\\[5pt]
Of course, the notion of separability extends easily to more than two sub-referee functions.
\subsection{Sampling process}
The definition~(\ref{DSmB3:2:F2K9:eq:def}) makes apparent a fusion process which is similar to a probabilistic conditional decision on the set of propositions.
Notice that the basic belief assignments are not related, in practice, to \emph{physical} probabilities.
But the implied mathematics are similar, as well as some concepts.
In particular, the fusion could be interpreted as a two stages process.
In a first stage, the sources of information generate independent entries according to the respective beliefs.
Then, a final decision is done by the referee function conditionally to the entries.
As a result, an output is produced or not.
\\[5pt]
This interpretation has two profitable consequences.
First at all, it provides an intuitive background for constructing new rules: in our framework, a new rule is just the design of a new referee.
Secondly, our interpretation makes possible sampling methods in order to approximate and accelerate complex fusion processes.
Notice that the sampling method is used here as a mathematical tool for approximating the \emph{belief}, not for simulating an individual choice.
Indeed, evidence approaches deal with belief on propositions, not with individual propositions.
\paragraph{Sampling algorithm.}
Samples of the fused basic belief assignment $\oplus[m_{1:s}|F]$ are generated by iterating the following processes:
\begin{description}
\item[]\rien\vspace{-10pt}
\begin{description}
\item[Entries generation:] For each $i\in[\![1,s]\!]$\,, generates $Y_i\in\genFrame\Theta$ according to the basic belief assignment $m_i$, considered as a probabilistic distribution over the set $\genFrame\Theta$\,,
\item[Conditional arbitrament:] \rien
\begin{enumerate}
\item Generate $X\in\genFrame\Theta$ according to referee function $F(X|Y_{1:s};m_{1:s})$, considered as a probabilistic distribution over the set $\genFrame\Theta$\,,
\item In the case $X=\emptyset$, reject the sample.
Otherwise, keep the sample.
\end{enumerate}
\end{description}
\end{description}
The performance of the sampling algorithm is at least dependent of two factors.
First at all, a fast implementation of the arbitrament is necessary.
Secondly, low rejection rate is better.
Notice however that the rejection rate is not a true handicap.
Indeed, high rejection rate means that the incident bbas are not compatible in regard to the fusion rule: these bba should not be fused.
By the way, the ratio of rejected samples will provide an empirical estimate of the rejection rate of the law.
\paragraph{The case of separable referee function.}
Assume that $F$ is separable into $G$ and $H$, \emph{i.e.} there are $\theta\in[0,1]$ and a partition $\left(\{u_{1:r}\},\{v_{1:t}\}\right)$ of $[\![1,s]\!]$ such that:
$$
F(X|Y_{1:s};m_{1:s})=
\theta G(X|Y_{u_{1:r}};m_{u_{1:r}})
 + 
(1-\theta) H(X|Y_{v_{1:t}};m_{v_{1:t}})\;.
$$
Then, samples of the fused basic belief assignment $\oplus[m_{1:s}|F]$ are generated by means of the sub-arbitraments related to $G$ and $H$:
\begin{description}
\item[]\rien\vspace{-10pt}
\begin{description}
\item[Choice of a sampling sub-process:]\rien
\begin{enumerate}
\item Generate a random number $x\in[0,1]$ according to the uniform distribution, 
\item If $x<\theta$ then jump to {\bf Sampling process related to $G$},
\item Otherwise jump to {\bf Sampling process related to $H$},
\end{enumerate}
\item[Sampling process related to $G$:]\rien
\begin{description}
\item[Entries generation:] For each $i\in[\![1,r]\!]$\,, generates $Y_{u_i}\in\genFrame\Theta$ according to the basic belief assignment $m_{u_i}$\,,
\item[Conditional arbitrament:] \rien
\begin{enumerate}
\item Generate $X\in\genFrame\Theta$ according to $G(X|Y_{u_{1:r}};m_{u_{1:r}})$\,,
\item In the case $X=\emptyset$, reject the sample.
Otherwise, return the sample.
\end{enumerate}
\end{description}
\item[Sampling process related to $H$:]\rien
\begin{description}
\item[Entries generation:] For each $i\in[\![1,t]\!]$\,, generates $Y_{v_i}\in\genFrame\Theta$ according to the basic belief assignment $m_{v_i}$\,,
\item[Conditional arbitrament:] \rien
\begin{enumerate}
\item Generate $X\in\genFrame\Theta$ according to $H(X|Y_{v_{1:t}};m_{v_{1:t}})$\,,
\item In the case $X=\emptyset$, reject the sample.
Otherwise, return the sample.
\end{enumerate}
\end{description}
\end{description}
\end{description}
This result is a direct consequence of the property of separability~(\ref{DSmB3:2:F2K9:eq:separable}).
This algorithm will spare the sampling of useless entries.
Therefore, it is more efficient to implement the separability when it is possible.
\subsection{Algorithmic definition of fusion rules}
As seen previously, fusion rules based on referee functions are easily approximated by means of sampling process.
This sampling process is double-staged.
The first stage computes the samples related to the entry bbas $m_{1:s}$\,.
The second stage computes the fused samples by a conditional arbitrament between the different hypotheses.
This arbitrament is formalized by a referee function.
\\[5pt]
In practice, it is noteworthy that there is no need for a mathematical definition of the referee function.
The only important point is to be able to compute the arbitrament.
We have here a new approach for defining fusion rules of evidences.
Fusion rules may be defined entirely by the means of an algorithm for computing the conditional arbitrament.
\paragraph{Assertion.}
\emph{There are three equivalent approaches for defining fusion rules in the paradigm of referee:
\begin{itemize}
\item By defining a formula which maps the entry bbas $m_{1:s}$ to the fused bba $m_1\oplus\cdots\oplus m_s$ (classical approach),
\item By defining a referee function $F$, which makes the conditional arbitrament $F(X|Y_{1:s};m_{1:s})$\,,
\item By constructing an algorithm which actually makes the conditional arbitrament between $Y_{1:s}$ in favor of $X$\,.  
\end{itemize}%
}\rien\\[-5pt]
It is sometime much easier and more powerful to just construct the algorithm for conditional arbitrament.
\\[5pt]
The following section illustrates the afore theoretical discussion on well known examples.
%
\section{Example of referee functions}
\label{DSmB3:2:F2K9:section:3}
Let be given $s$ sources of information characterized by their bbas $m_{1:s}$.
\subsection{Dempster-shafer rule}

The fused bba $m_{\text{\tiny{DST}}}$ obtained from $m_{1:s}$ by means of \emph{Dempster-Shafer} fusion rule is defined by equation~(\ref{DSmB3:2:eqfond:1}).
It has an immediate interpretation by means of referee functions.
\paragraph{Definition by referee function.}
The definition of a referee function for Dempster-Shafer is immediate:
$$
m_{\text{\tiny{DST}}}=\oplus[m_{1:s}|F_{\wedge}]
\mbox{ \ where }
F_{\wedge}(X|Y_{1:s};m_{1:s}) = I\left[X=\bigcap_{k=1}^{s} Y_{k}\right]
 \;.
$$
\paragraph{Algorithmic definition.}
The algorithmic implementation of $F_{\wedge}$ is described subsequently and typically implies possible conditional rejections: 
\\\begin{sloppypar}
{\bf Conditional arbitrament:}
\begin{enumerate}
\item Set $X=\bigcap_{k=1}^{s} Y_k$\;, 
\item If $X =\emptyset$\,, then reject the sample. Otherwise, keep the sample.
\end{enumerate}
\end{sloppypar}
\subsection{Disjunctive rule}
The fused bba $m_{\vee}$ obtained from $m_{1:s}$ by means of the disjunctive rule is defined by equation~(\ref{DSmB3:2:eqfond:2:b}).
It has an immediate interpretation by means of referee functions.
\paragraph{Definition by referee function.}
The definition of a referee function for the disjunctive rule is immediate from~(\ref{DSmB3:2:eqfond:2:b}):
$$
m_{\vee}=\oplus[m_{1:s}|F_{\vee}]
$$
where:
$$
F_{\vee}(X|Y_{1:s};m_{1:s}) =
 I\left[X=\bigcup_{k=1}^{s} Y_{k}\right]\;.
$$
\paragraph{Algorithmic definition.}
The algorithmic implementation of $F_{\vee}$ is described subsequently.
It does not imply rejections: 
\\\begin{sloppypar}
{\bf Conditional arbitrament:}
\begin{enumerate}
\item Set $X=\bigcup_{k=1}^{s} Y_k$\;. 
\end{enumerate}
\end{sloppypar}
\subsection{Dubois $\&$ Prade rule}
The fused bba $m_{\text{\tiny{D\&P}}}$ obtained from $m_{1:s}$ by means of the rule of Dubois $\&$ Prade is defined by equation~(\ref{DSmB3:2:eqfond:3}).
It has an interpretation by means of referee functions.
\paragraph{Definition by referee function.}
The definition of a referee function for Dubois $\&$ Prade rule is deduced from~(\ref{DSmB3:2:eqfond:3}):
$$
m_{\text{\tiny{D\&P}}}=\oplus[m_{1:s}|F_{\text{\tiny{D\&P}}}]
$$
where:
$$\begin{array}{@{}l@{}}\displaystyle
F_{\text{\tiny{D\&P}}}(X|Y_{1:s};m_{1:s})
 \vspace{5pt}\\\displaystyle
 \rien\qquad\qquad=
 I\left[\bigcap_{k=1}^{s} Y_{k}\ne\emptyset\right]F_{\wedge}(X|Y_{1:s};m_{1:s})
 +
 I\left[\bigcap_{k=1}^{s} Y_{k}=\emptyset\right] F_{\vee}(X|Y_{1:s};m_{1:s})
 \vspace{5pt}\\\displaystyle
 \rien\qquad\qquad=
 I\left[X=\bigcap_{k=1}^{s} Y_{k}\ne\emptyset\right]
 +
 I\left[\bigcap_{k=1}^{s} Y_{k}=\emptyset\right] I\left[X=\bigcup_{k=1}^{s} Y_{k}\right]
  \;.
\end{array}$$
The first formulation of $F_{\text{\tiny{D\&P}}}$ is particularly interesting, since it illustrates how to construct a referee function by means of a conditional branching to already existing referee functions.
In the case of Dubois $\&$ Prade rule, the rule has a disjunctive behavior when there is a conjunctive conflict, \emph{i.e.} $\bigcap_{k=1}^{s} Y_{k}=\emptyset$, and a conjunctive behavior otherwise.
Thus, the referee function is obtained as the summation of the exclusive sub-arbitraments: 
$$
I\left[\bigcap_{k=1}^{s} Y_{k}=\emptyset\right] F_{\vee}(X|Y_{1:s};m_{1:s})\quad\mbox{(disjunctive case)}
$$
and
$$
I\left[\bigcap_{k=1}^{s} Y_{k}\ne\emptyset\right]F_{\wedge}(X|Y_{1:s};m_{1:s})\quad\mbox{(conjunctive case).}
$$
\paragraph{Algorithmic definition.}
The algorithmic implementation of $F_{\text{\tiny{D\&P}}}$ is described subsequently.
It does not imply rejections: 
\\\begin{sloppypar}
{\bf Conditional arbitrament:}
\begin{enumerate}
\item Set $X=\bigcap_{k=1}^{s} Y_k$\;, 
\item If $X =\emptyset$\,, then set $X=\bigcup_{k=1}^{s} Y_k$\;.
\end{enumerate}
\end{sloppypar}
\subsection{Averaging rule}
Let be given the averaging parameters $\alpha_{1:s}\ge0$ such that $\sum_{i=1}^s\alpha_i=1$\,.
The averaged bba $m_{\mu}[\alpha]=\sum_{i=1}^s \alpha_i m_i$ could be obtained by means of a refereee function.
\paragraph{Averaging by referee function.}
The definition of a referee function for averaging is immediate:
$$
m_{\mu}[\alpha]=\oplus[m_{1:s}|F_{\mu}[\alpha]]
$$
where:
$$
F_{\mu}[\alpha](X|Y_{1:s};m_{1:s})=
 \sum_{i=1}^s \alpha_i\;I\left[X=Y_{i}\right]
\;.
$$
\paragraph{Proof by applying the separability.}
It is noticed that $F$ is separable:
$$
F_{\mu}[\alpha](X|Y_{1:s};m_{1:s})=
 \sum_{i=1}^s \alpha_i\;\mathrm{id}_i(X|Y_i,m_i)
\;,
$$
where the referee function $\mathrm{id}_i$ is defined by:
$$
\mathrm{id}_i(X|Y_i,m_i)=I[X=Y_i]\quad\mbox{for any }X,Y_i\in\genFrame\Theta
$$
By applying the separability property~(\ref{DSmB3:2:F2K9:eq:separable}), $s-1$ times, it comes:
$$
\oplus[m_{1:s}|F_{\mu}[\alpha]]=\sum_{i=1}^s \alpha_i\;\oplus[m_{i}|\mathrm{id}_i]\;.
$$
It happens that:
$$
\oplus[m_{i}|\mathrm{id}_i]=m_i\;,
$$
so that:
$$
\oplus[m_{1:s}|F_{\mu}[\alpha]]=\sum_{i=1}^s \alpha_i\;m_{i}=m_{\mu}[\alpha]\;.
$$
\paragraph{Algorithmic implementation of averaging.}
It is interesting here to implement the separability of the referee function.
Therefore, the entire sampling algorithm is derived from the separability property:
\begin{description}
\item[]\rien\vspace{-10pt}
\begin{description}
\item[Choice of a sampling sub-process:]\rien
\begin{enumerate}
\item Generate a random integer $i\in[\![1, s]\!]$ according to probability $\alpha$\,,
\item Jump to {\bf Sampling process related to $\mathrm{id}_i$},
\end{enumerate}
\item[Sampling process related to $\mathrm{id}_i$:]\rien
\begin{description}
\item[Entries generation:] Generates $Y_i\in\genFrame\Theta$ according to the basic belief assignment $m_i$\,,
\item[Conditional arbitrament:]
Return $X=Y_i$\;.
\end{description}
\end{description}
\end{description}
\subsection{PCR6 rule}
The fused bba $m_{\text{\tiny{PCR6}}}$ obtained from $m_{1:s}$ by means of the PCR6 rule is defined by equation~(\ref{DSmB3:2:eqfond:4}).
It has an interpretation by means of referee functions.
\paragraph{Definition by referee function.}
Definition~(\ref{DSmB3:2:eqfond:4}) could be reformulated into:
$$
m_{\text{\tiny{PCR6}}}(X)=m_{\wedge}(X) 
+
\sum_{i=1}^s \sum_{\substack{
\bigcap_{k=1}^{s} Y_{k} = \emptyset
 \\
 Y_1,\cdots,Y_s\in \genFrame\Theta
 }}
 \left(\frac{\displaystyle
I[X=Y_i]\;m_i(Y_i)\prod_{j=1}^{s}
m_{j}(Y_{j}) 
 }{\displaystyle
\sum_{j=1}^{s}
m_{j}(Y_j) 
 }\right)\;,
$$
and then:
\begin{equation}
m_{\text{\tiny{PCR6}}}(X)=m_{\wedge}(X) 
+
\sum_{\substack{
\bigcap_{k=1}^{s} Y_{k} = \emptyset
 \\
 Y_1,\cdots,Y_s\in \genFrame\Theta
 }}
 \prod_{i=1}^{s}
m_{i}(Y_{i}) 
\ \frac{\displaystyle
\sum_{j=1}^s I[X=Y_j]\;m_j(Y_j)
 }{\displaystyle
\sum_{j=1}^{s}
m_{j}(Y_j) 
 }
 \;.
 \label{eq:mPCR6:3}
\end{equation}
At last, it is derived a formulation of PCR6 by means of a referee function:
$$
m_{\text{\tiny{PCR6}}}=
\oplus[m_{1:s}|F_{\text{\tiny{PCR6}}}]\;,
$$
where the referee function $F_{\text{\tiny{PCR6}}}$ is defined by:
\begin{multline}
F_{\text{\tiny{PCR6}}}(X|Y_{1:s};m_{1:s})=
\\
I\left[X=\bigcap_{k=1}^{s} Y_{k} \ne\emptyset\right]
+I\left[\bigcap_{k=1}^{s} Y_{k} = \emptyset\right]
\ \frac{\displaystyle\sum_{j=1}^s I[X=Y_j]\;m_j(Y_j)
 }{\displaystyle
\sum_{j=1}^{s}
m_{j}(Y_j) 
 }
 \;.
 \label{eq:mPCR6:4}
\end{multline}
\paragraph{Algorithmic definition.}
Again, the algorithmic implementation is immediate:
\\\begin{sloppypar}
{\bf Conditional arbitrament:}
\begin{enumerate}
\item If $\bigcap_{k=1}^{s} Y_{k} \ne\emptyset$\,, then set $X=\bigcap_{k=1}^{s} Y_{k}$
\item Otherwise:
\begin{enumerate}
\item Define the probability $P$ over $[\![1,s]\!]$ by:
$$
P_i=\frac{m_{i}(Y_i)}{\sum_{j=1}^{s}
m_{j}(Y_j)}\mbox{ \ for any }i\in[\![1,s]\!]\;,
$$
\item Generate a random integer $k\in[\![1,s]\!]$ according to $P$\,,
\item set $X=Y_k$\;. 
\end{enumerate}
\end{enumerate}
\end{sloppypar}\rien\\
It is noticed that this process does not produce any rejection case $X=\emptyset$.
As a consequence, the last rejection step has been removed.
\\[5pt]
Essentially, this algorithm distinguishes two cases:
\begin{itemize}
\item there is a consensus; then, answer the consensus,
\item there is not a consensus; then choose an entry among all entries proportionally to its belief.
It is noteworthy that there is no attempt to transform the entries in this case.
\end{itemize}
This algorithm is efficient and is not time-consuming.
The whole sampling approach should be a good alternative for approximating PCR6, particularly on large frames of discernment.
\subsection{Non conjunctive rejection}
As seen in section~\ref{DSmB3:2:F2K9:section:2}, the rejection, resulting from a fusion based on referee functions, is not necessary a conjunctive conflict: this rejection is the result of the arbitrament rejections $F(\emptyset|Y_{1:s};m_{1:s})$\,.
An example of rule with disjunctive rejection is proposed now.
This example is rather unnatural; is is only constructed for illustration. 
The context of fusion is the following:
\begin{description}
\item[Case a:] Entries which union is $\Omega$ (\emph{i.e.} is totally imprecise) are rejected; the idea here, is to reject entries which are insufficiently focused,
\item[Case b:] An averaging of the bba is done, otherwise.
\end{description}
The referee function is thus obtained by means of a conditional branching to the rejection (case a) or to the averaging (case b).
As a consequence, the referee function is obtained as the summation of the following exclusive sub-arbitraments: 
\begin{description}
\item[Case a:] $\displaystyle
I\left[\bigcup_{i=1}^{s} Y_{i}=\Omega\right]I\left[X=\emptyset\right]\;,
$
\item[Case b:] $\displaystyle
I\left[\bigcup_{i=1}^{s} Y_{i}\ne\Omega\right]\sum_{i=1}^s\frac{I\left[X=Y_i\right]}{s}\;.
$
\end{description}
The referee function for this rule is:
$$
F(X|Y_{1:s};m_{1:s})=
I\left[\bigcup_{i=1}^{s} Y_{i}=\Omega\right]I\left[X=\emptyset\right]
+
I\left[\bigcup_{i=1}^{s} Y_{i}\ne\Omega\right]\sum_{i=1}^s\frac{I\left[X=Y_i\right]}{s}\;.
$$
Let us apply this rule to the example of $s=2$ bbas $m_1$ and $m_2$ defined on $2^{\{a,b,c\}}$ by $m_1(\{a\})=0.1$\,,  $m_1(\{a,b\})=0.9$\,, $m_2(\{b\})=0.2$ and $m_2(\{b,c\})=0.8$\,.
Then, the fused bba $m=m_1\oplus m_2[F]$ is obtained as:
$$\begin{array}{@{}l@{}}
z=m_1(\{a\})m_2(\{b,c\})+m_1(\{a,b\})m_2(\{b,c\})=0.8\;,\\[5pt]
m(\{a\})=\frac{m_1(\{a\})m_2(\{b\})}{2(1-z)}=0.05\;,\quad 
m(\{a,b\})=\frac{m_1(\{a,b\})m_2(\{b\})}{2(1-z)}=0.45\;,\\[5pt]
m(\{b\})=\frac{m_1(\{a\})m_2(\{b\})+m_1(\{a,b\})m_2(\{b\})}{2(1-z)}=0.5\;.
\end{array}$$
In this example, the rejection rate $z$ is not conjunctive, since it involves $\{a,b\}$ and $\{b,c\}$ such that $\{a,b\}\cap\{b,c\}\ne\emptyset$.
\subsection{Any rule?}
\emph{Is it possible to construct a referee function for any existing fusion rule?}
\\[5pt]
Actually, the answer to this question is ambiguous.
If it is authorized that $F$ depends on $m_{1:s}$ without restriction, then the theoretical answer is trivially yes.
\paragraph{Property.}
Let be given the fusion rule $m_1\oplus\cdots\oplus m_s$\,, applying on the bbas $m_{1:s}$.
Define the referee function $F$ by:
$$
F(X|Y_{1:s};m_{1:s})=m_1\oplus\cdots\oplus m_s(X)\mbox{ \ for any }X,Y_{1:s}\in\genFrame\Theta\;.
$$
Then $F$ is actually a referee function and $\oplus[m_{1:s}|F]=m_1\oplus\cdots\oplus m_s$\,.
\\[5pt]
\emph{Proof is immediate.}
\\[5pt]
Of course, this result is useless in practice, since such referee function is inefficient.
It is inefficient because it does not provide an intuitive interpretation of the rule, and is as difficult to compute as the fusion rule.
Then, it is useless to have a sampling approach with such definition.
\\[5pt]
As a conclusion, referee functions have to be considered together with their efficiency.
The efficiency of referee function is not a topic which is studied in this chapter.

\section{A new rule: PCR$\sharp$}
\label{DSmB3:2:F2K9:section:4}
\paragraph{Definition.}
For any $k\in[\![1,s]\!]$, it is defined:
$$
C[k|s]=\left\{\gamma\subset[\![1,s]\!]\left/\mathrm{card}(\gamma)=k\right.\right\}\;,
$$ 
the set of $k$-combinations of $[\![1,s]\!]$\,. 
Of course, the cardinal of $C[k|s]$ is $\left(\begin{array}{@{}c@{}}s\\k\end{array}\right)$\,.
\\[5pt]
For convenience, the undefined object $C[s+1|s]$ is actually defined by:
$$
C[s+1|s]=\bigl\{\{\emptyset\}\bigr\}\;,
$$
so as to ensure:
$$
\min_{\gamma\in C[s+1|s]}
\left\{
I\left[\bigcap_{i\in\gamma}Y_i=\emptyset\right]
\right\}=1
$$
\subsection{Limitations of PCR6}
The algorithmic interpretation of PCR6 has shown that PCR6 distinguishes two cases:
\begin{itemize}
\item The entry information are compatible; then, the conjunctive consensus is decided,
\item The entry information are not compatible; then, a mean decision is decided, weighted by the relative beliefs of the entries.
\end{itemize}
In other words, PCR6 only considers consensus or no-consensus cases.
But for more than $2$ sources, there are many cases of \emph{intermediate consensus}.
By construction, PCR6 is not capable to manage intermediate consensus.
This is a notable limitation of PCR6.
\\[5pt]
The new rule PCR$\sharp$, which is defined now, extends PCR6 by considering partial consensus in addition to full consensus and absence of consensus.
This rule is constructed by specifying the arbitrament algorithm.
Then, a referee function is deduced.
\subsection{Algorithm}
The following algorithm tries to reach a maximal consensus.
It first tries the full consensus, then consensus of $s-1$ sources, $s-2$ sources, and so on, until a consensus is finally found.
When several consensus with $k$ sources is possible, the final answer is chosen randomly, proportionally to the beliefs of the consensus.
In the following algorithm, comments are included preceded by $//$ (c++ convention).\\
\begin{sloppypar}
{\bf Conditional arbitrament:}
\begin{enumerate}
\item Set $stop={\tt false}$ and $k=s$\,,
\\\emph{$//$ $k$ is the size of the consensus, which are searched. At beginning, it is maximal.}
\item\label{DSmB3:2:algoSharp:1}
For each $\gamma\in C[k|s]$\,, do:
\\\emph{$//$ All possible consensus of size $k$ are tested.}
\begin{enumerate}
\item If $\bigcap_{i\in\gamma}Y_i\ne\emptyset$\;, then set $\omega_\gamma=\prod_{i\in\gamma}m_i(Y_i)$ and $stop={\tt true}$\;,
\\\emph{$//$ If a consensus of size $k$ is found to be functional, then it is no more necessary to diminish the size of the consensus.
This is done by changing the value of Boolean $stop$.}
\\\emph{$//$ Moreover, the functional consensus are weighted by their beliefs.} 
\item Otherwise set $\omega_\gamma=0$\;,
\\\emph{$//$ Non-functional consensus are weighted zero.} 
\end{enumerate}	
\item If $stop={\tt false}$\,, then set $k=k-1$ and go back to~\ref{DSmB3:2:algoSharp:1}\,,
\\\emph{$//$ If no functional consensus of size $k$ has been found, then it is necessary to test smaller sized consensus.
The process is thus repeated for size $k-1$.}
\item Choose $\gamma\in C[k|s]$ randomly, according to the probability:
 $$
P_\gamma=\frac{\omega_\gamma}{\sum_{\gamma\in C[k|s]}\omega_\gamma}\;,
 $$
\\\emph{$//$ Otherwise, choose a functional consensus.
Here, the decision is random and proportional to the consensus belief.}
\item At last, set $X=\bigcap_{i\in \gamma}Y_i\;.$
\\\emph{$//$ Publish the sample related to the consensus.}
\end{enumerate}	
\end{sloppypar}
\paragraph{Algorithm without comment.}\rien\\[5pt]
\begin{sloppypar}
{\bf Conditional arbitrament:}
\begin{enumerate}
\item Set $stop={\tt false}$ and $k=s$\,,
\item\label{DSmB3:2:algoSharp:1b}
For each $\gamma\in C[k|s]$\,, do:
\begin{enumerate}
\item If $\bigcap_{i\in\gamma}Y_i\ne\emptyset$\;, then set $\omega_\gamma=\prod_{i\in\gamma}m_i(Y_i)$ and $stop={\tt true}$\;,
\item Otherwise set $\omega_\gamma=0$\;,
\end{enumerate}	
\item If $stop={\tt false}$\,, then set $k=k-1$ and go back to~\ref{DSmB3:2:algoSharp:1b}\,,
\item Choose $\gamma\in C[k|s]$ randomly, according to the probability:
 $$
 P_\gamma=\frac{\omega_\gamma}{\sum_{\gamma\in C[k|s]}\omega_\gamma}\;,
 $$
\item At last, set $X=\bigcap_{i\in \gamma}Y_i\;.$
\end{enumerate}	
\end{sloppypar}
\subsection{Referee function}
Historically, PCR$\sharp$ has been defined by means of an algorithm, not by means of a formal definition of the referee function.
It is however possible to give a formal definition of the referee function which is equivalent to the algorithm:
\begin{equation}\label{DSmB3:2:Sharp:eq:1}\mbox{\small$\begin{array}{@{}l}
\displaystyle
F_{\text{\tiny{PCR$\sharp$}}}(X|Y_{1:s};m_{1:s})=
\sum_{k=1}^s
\ \min_{
\gamma\in C[k+1|s]
}
\left\{
I\left[\bigcap_{i\in\gamma}Y_i=\emptyset\right]
\right\}
\vspace{5pt}\\\displaystyle
\rien\qquad\times\min\left\{\max_{\gamma\in C[k|s]}\left\{
I\left[\bigcap_{i\in\gamma}Y_i\ne\emptyset\right]
\right\},
\frac{\displaystyle
\sum_{\gamma\in C[k|s]}
I\left[X=\bigcap_{i\in\gamma}Y_i\ne\emptyset\right]\prod_{i\in\gamma}m_i(Y_i)
}{\displaystyle
\sum_{\gamma\in C[k|s]}
I\left[\bigcap_{i\in\gamma}Y_i\ne\emptyset\right]\prod_{i\in\gamma}m_i(Y_i)
}
\right\}\;.
\end{array}$}\end{equation}
\emph{Sketch of the proof.}
The following correspondences are established between the arbitrament algorithm and the referee function:
\begin{itemize}
\item The summation $\sum_{k=1}^s$ is a formalization of the loop from $k=s$ down to $k=1$\,,
\item At step $k$, the component:
$$
\min_{
\gamma\in C[k+1|s]
}
\left\{
I\left[\bigcap_{i\in\gamma}Y_i=\emptyset\right]
\right\}
$$
ensures that there is not a functional consensus of larger size $j>k$\,.
Typically, the component is $0$ if a larger sized functional consensus exists, and $1$ otherwise.
This component is complementary to the summation, as it formalizes the end of the loop, when a functional consensus is actually found,
\item At step $k$, the component:
$$
\Omega=\frac{\displaystyle
\sum_{\gamma\in C[k|s]}
I\left[X=\bigcap_{i\in\gamma}Y_i\ne\emptyset\right]\prod_{i\in\gamma}m_i(Y_i)
}{\displaystyle
\sum_{\gamma\in C[k|s]}
I\left[\bigcap_{i\in\gamma}Y_i\ne\emptyset\right]\prod_{i\in\gamma}m_i(Y_i)
}
$$
encodes the choice of a functional consensus of size $k$, proportionally to its belief.
The chosen consensus results in the production of the sample $X$\,, 
\item At step $k$, the component:
$$
\max_{\gamma\in C[k|s]}\left\{
I\left[\bigcap_{i\in\gamma}Y_i\ne\emptyset\right]
\right\}
$$
tests if there is a functional consensus of size $k$.
The component answers $1$ if such consensus exists, and $0$ otherwise.
It is combined with a minimization of the form:
$$
\min\left\{\max_{\gamma\in C[k|s]}\left\{
I\left[\bigcap_{i\in\gamma}Y_i\ne\emptyset\right]
\right\},
\Omega\right\}\;,\mbox{ \ where }\Omega\le1\;.
$$
This is some kind of ``{\tt if ... then}''\ :
if a functional consensus of size $k$ exists, then the value $\Omega$ is computed.
Otherwise, it is the value $0$\,.
Since the value $\Omega$ encodes a sampling decision, we have here sampling decision, which is conditioned by the fact that a functional consensus exists.
\end{itemize}
The equivalence is a consequence of these correspondences.
\\$\Box$
\subsection{Variants of PCR$\sharp$}
Actually, $\mathrm{card}(C[k|s])=\left(\begin{array}{@{}c@{}}s\\k\end{array}\right)$ increases quickly when $s$ is great and $k$ is not near $1$ or $s$\,. 
As a consequence, PCR$\sharp$ implies hard combinatorics, when used in its general form.
On the other hand, it may be interesting to reject samples, when a consensus is not possible with a minimal quorum.
In order to address such problems, a slight extension of PCR$\sharp$ is proposed now.
\paragraph{Algorithm.}\rien\\[5pt]
Let $r\in[\![1,s]\!]$ and let $k_{1:r}\in[\![1,s]\!]$ be a decreasing sequence such that:
$$
s\ge k_1>\cdots> k_r\ge1\;.
$$
For convenience, the undefined object $k_0$ is actually defined by:
$$
k_0=s+1\;,
$$
so as to ensure:
$$
\min_{\gamma\in C[k_0|s]}
\left\{
I\left[\bigcap_{i\in\gamma}Y_i=\emptyset\right]
\right\}=1
$$
Then, the rule PCR$\sharp[k_{1:r}]$ is defined by the following algorithm.
\\[5pt]
\begin{sloppypar}
{\bf Conditional arbitrament:}
\begin{enumerate}
\item Set $stop={\tt false}$ and $h=1$\,,
\item\label{DSmB3:2:algoSharp:2}
For each $\gamma\in C[k_h|s]$\,, do:
\begin{enumerate}
\item If $\bigcap_{i\in\gamma}Y_i\ne\emptyset$\;, then set $\omega_\gamma=\prod_{i\in\gamma}m_i(Y_i)$ and $stop={\tt true}$\;,
\item Otherwise set $\omega_\gamma=0$\;,
\end{enumerate}	
\item If $stop={\tt false}$\,, then:
\begin{enumerate}
\item set $h=h+1$\,,
\item If $h\le r$, go back to~\ref{DSmB3:2:algoSharp:2}\,,
\end{enumerate}
\item If $h>r$, then reject the entries and {\bf end}, 
\item Otherwise, choose $\gamma\in C[k_h|s]$ randomly, according to the probability:
 $$
 P_\gamma=\frac{\omega_\gamma}{\sum_{\gamma\in C[k_h|s]}\omega_\gamma}\;,
 $$
\item Set $X=\bigcap_{i\in \gamma}Y_i\;.$ and {\bf end}.
\end{enumerate}	
\end{sloppypar}
\paragraph{Referee function}
$$\mbox{\small$\begin{array}{@{}l}
\displaystyle
F_{\mbox{{\tiny PCR$\sharp[k_{1:r}]$}}}(X|Y_{1:s};m_{1:s})=
\vspace{5pt}\\\displaystyle
\rien\qquad
I[X=\emptyset]\min_{
\gamma\in C[k_r|s]
}
\left\{
I\left[\bigcap_{i\in\gamma}Y_i=\emptyset\right]
\right\}
+
\sum_{h=1}^r
\ \min_{
\gamma\in C[k_{h-1}|s]
}
\left\{
I\left[\bigcap_{i\in\gamma}Y_i=\emptyset\right]
\right\}
\vspace{5pt}\\\displaystyle
\rien\qquad\times\min\left\{\max_{\gamma\in C[k_h|s]}\left\{
I\left[\bigcap_{i\in\gamma}Y_i\ne\emptyset\right]
\right\},
\frac{\displaystyle
\sum_{\gamma\in C[k_h|s]}
I\left[X=\bigcap_{i\in\gamma}Y_i\ne\emptyset\right]\prod_{i\in\gamma}m_i(Y_i)
}{\displaystyle
\sum_{\gamma\in C[k_h|s]}
I\left[\bigcap_{i\in\gamma}Y_i\ne\emptyset\right]\prod_{i\in\gamma}m_i(Y_i)
}
\right\}\;.
\end{array}$}$$
\emph{proof is left to the reader.}
\paragraph{PCR6 and PCR$\sharp$.}
Assume that PCR6 is applied to $s$ entries $m_{1:s}$\,.
Then:
\begin{center}
PCR6=PCR$\sharp[s,1]$
\end{center}
\paragraph{DST and PCR$\sharp$.}
Assume that DST is applied to $s$ entries $m_{1:s}$\,.
Then:
\begin{center}
DST=PCR$\sharp[s]$
\end{center}
\paragraph{Variant with truncation and rejection.}
Let $r\in[\![1,s]\!]$\,.
The rule PCR$\sharp[s,s-1,\cdots,r]$ will search for maximally sized functional consensus.
If it is not possible to find functional consensus with size greater or equal to $r$\,, the algorithm rejects the entries.
\paragraph{Variant with truncation and final mean decision.}
Let $r\in[\![1,s]\!]$\,.
The rule PCR$\sharp[s,s-1,\cdots,r,1]$ will search for maximally sized functional consensus.
If it is not possible to find functional consensus with size greater or equal to $r$\,, the algorithm choose an entry proportionally to its belief.
%
%
\section{Numerical examples}
\label{DSmB3:2:f2k9:F2K9:sect:Numexample}
It is assumed:
$$
\genFrame\Theta=\left\{\emptyset,\{a\},\{b\},\{c\},\{b,c\},\{c,a\},\{a,b\},\{a,b,c\}\right\}\;.
$$
Various examples of bbas, $m_{1:s}$\,, are considered on $\genFrame\Theta$ and fused by means of rules based on different referee functions.
The fused rule, $m=\oplus[m_{1:s}|F]$, is computed both mathematically or by means of the sampler.
When the fusion is obtained by sampling a particle cloud, the fused bba estimate is deduced from an empirical averaging.
The complete process arises as follows:
\begin{enumerate}
\item Repeat from $n=1$ to $n=N$:
\begin{enumerate}
\item Generate the particle $X_{n}\in\genFrame\Theta$ by sampling $\oplus[m_{1:s}|F]$\,,
\item If the sampling process failed, then set $X_{n}=\mathrm{rejected}$\,,
\end{enumerate}
\item Compute $\widehat{z}$, the estimate of $z$, by setting $\displaystyle\widehat{z}=\frac1N\sum_{n=1}^NI[X_n=\mathrm{rejected}]$\,,
\item For any $X\in\genFrame\Theta$, compute $\widehat{m}(X)$, the estimate of $m(X)$, by:
$$
\widehat{m}(X)=\frac1{N(1-\widehat{z})} \sum_{n=1}^N I[X_n=X]\;.
$$
\end{enumerate} 
It is known that the accuracy of this estimate is of the magnitude of $\frac1{\sqrt{N}}$\,.
Typically, when the rejection rate is zero, \emph{i.e.} $z=0$, the variance $\sigma(\widehat{m}(X))$ is given by:
$$
\sigma(m(X))=\sqrt{\frac{m(X)\cdot(1-m(X))}{N}}\;.
$$
\subsection{Convergence}
\paragraph{Example 1.}
The bbas $m_1$ and $m_2$ are defined by:
$$\begin{array}{@{}l@{}}\displaystyle
m_1(\{a,b\})=0.2\,,\ 
m_1(\{a,c\})=0.4\,,\ 
m_1(\{b,c\})=0.3\,,\ 
m_1(\{a,b,c\})=0.1\,,
\vspace{4pt}\\\displaystyle
m_2(\{a,b\})=0.4\,,\ 
m_2(\{a,c\})=0.2\,,\ 
m_2(\{b,c\})=0.3\,,\ 
m_2(\{a,b,c\})=0.1\,.
\end{array}$$
These bbas are fused by means of DST, resulting in $m=m_{\mbox{\tiny DST}}$:
\begin{center}
$z=0$,
$m(\{a\})=0.2$,
$m(\{b\})=0.18$,
$m(\{a,b\})=0.14$,
$m(\{c\})=0.18$,
$m(\{a,c\})=0.14$,
$m(\{b,c\})=0.15$,
$m(\{a,b,c\})=0.01$.
\end{center}
The estimate $\widehat{m}$ of $m$ is obtained by the following process:
\begin{enumerate}
\item Repeat from $n=1$ to =$n=N$:
\begin{enumerate}
\item Generate $Y_1$ and $Y_2$ by means of $m_1$ and $m_2$ respectively,
\item If $Y_1\cap Y_2=\emptyset$\,, then set $X_n=\mathrm{rejected}$\,,
\item Otherwise, set $X_n=Y_1\cap Y_2$\,,
\end{enumerate}
\item Set $\displaystyle\widehat{z}=\frac1N\sum_{n=1}^NI[X_n=\mathrm{rejected}]$\,,
\item For any $X\in\genFrame\Theta$\,, compute $\widehat{m}(X)$ by:
$$
\widehat{m}(X)=\frac1{N(1-\widehat{z})} \sum_{n=1}^N I[X_n=X]\;.
$$
\end{enumerate}
The following table compares the empirical estimates of $m$, computed by means of particle clouds of different sizes $N$:
$$\begin{array}{@{}l||l|l|l|l|l|l|l|l||l@{}}
\log_{10}N&1&2&3&4&5&6&7&8&\infty
\\\hline
\widehat{z}&0&0&0&0&0&0&0&0 &0
\\
\widehat{m}(\{a\})&0.2&0.18&0.202&0.201&0.200&0.200&0.200&0.200 &0.2
\\
\widehat{m}(\{b\})&0.1&0.19&0.173&0.182&0.181&0.180&0.180&0.180 &0.18
\\
\widehat{m}(\{a,b\})&0.3&0.14&0.139&0.138&0.139&0.140&0.140&0.140 &0.14
\\
\widehat{m}(\{c\})&0.1&0.15&0.179&0.177&0.181&0.180&0.180&0.180 &0.18
\\
\widehat{m}(\{a,c\})&0.3&0.17&0.141&0.136&0.140&0.140&0.140&0.140 &0.14
\\
\widehat{m}(\{b,c\})&0&0.15&0.153&0.155&0.149&0.150&0.150&0.150 &0.15
\\
\widehat{m}(\{a,b,c\})&0&0.02&0.013&0.011&0.010&0.010&0.010&0.010 &0.01
\end{array}$$
For this choice of $m_1$ and $m_2$, there is no conflict.
The theoretical accuracy defined previously thus applies.
The results are compliant with the theoretical accuracy.
\paragraph{Example 2.}
This is an example with reject.
The bbas $m_1$ and $m_2$ are defined by:
$$\begin{array}{@{}l@{}}\displaystyle
m_1(\{a\})=0.4\,,\ 
m_1(\{a,b\})=0.5\,,\ 
m_1(\{a,b,c\})=0.1\,,
\vspace{4pt}\\\displaystyle
m_2(\{c\})=0.4\,,\ 
m_2(\{b,c\})=0.5\,,\ 
m_2(\{a,b,c\})=0.1\,.
\end{array}$$
These bbas are fused by means of DST, resulting in $m=m_{\mbox{\tiny DST}}$:
\begin{center}
$z=0.56$,
$m(\{a\})=0.091$,
$m(\{b\})=0.568$,
$m(\{a,b\})=0.114$,
$m(\{c\})=0.091$,
$m(\{b,c\})=0.114$,
$m(\{a,b,c\})=0.022$.
\end{center}
The estimate $\widehat{m}$ of $m$ is obtained by the same process as for example 1. 
The following table compares the empirical estimates of $m$, computed by means of particle clouds of different sizes $N$:
$$\begin{array}{@{}l||l|l|l|l|l|l|l|l||l@{}}
\log_{10}N&1&2&3&4&5&6&7&8&\infty
\\\hline
\widehat{z}&0.7&0.6&0.56&0.558&0.560&0.559&0.560&0.560 &0.56
\\
\widehat{m}(\{a\})&0.667&0.15&0.109&0.088&0.092&0.090&0.091&0.091 &0.091
\\
\widehat{m}(\{b\})&0&0.375&0.573&0.565&0.566&0.569&0.568&0.568 &0.568
\\
\widehat{m}(\{a,b\})&0&0.125&0.107&0.114&0.114&0.114&0.114&0.114 &0.114
\\
\widehat{m}(\{c\})&0.333&0.225&0.091&0.091&0.091&0.091&0.091&0.091 &0.091
\\
\widehat{m}(\{b,c\})&0&0.125&0.107&0.125&0.115&0.114&0.114&0.114 &0.114
\\
\widehat{m}(\{a,b,c\})&0&0&0.013&0.017&0.022&0.022&0.022&0.022 &0.022
\end{array}$$
Notice that the theoretical accuracy should be corrected, since $z>0$\,.
\paragraph{Example 3.}
The bbas $m_1$ and $m_2$ are defined by:
$$\begin{array}{@{}l@{}}\displaystyle
m_1(\{a\})=0.5\,,\ 
m_1(\{a,b\})=0.1\,,\ 
m_1(\{a,b,c\})=0.4\,,
\vspace{4pt}\\\displaystyle
m_2(\{c\})=0.3\,,\ 
m_2(\{a,c\})=0.3\,,\ 
m_2(\{a,b,c\})=0.4\,.
\end{array}$$
These bbas are fused by means of PCR6, resulting in $m=m_{\mbox{\tiny PCR6}}$:
\begin{center}
$m(\{a\})=0.385$,
$m(\{b\})=0.04$,
$m(\{a,b\})=0.007$,
$m(\{c\})=0.199$,
$m(\{a,c\})=0.12$,
$m(\{b,c\})=0.249$.
\end{center}
It is noticed that $z=0$ in this case of PCR6.
Then, the estimate $\widehat{m}$ is obtained by the following process:
\begin{enumerate}
\item Repeat from $n=1$ to =$n=N$:
\begin{enumerate}
\item Generate $Y_1$ and $Y_2$ by means of $m_1$ and $m_2$ respectively,
\item If $Y_1\cap Y_2\ne\emptyset$\,, then set $X_n=Y_1\cap Y_2$\,,
\item Otherwise, do:
\begin{enumerate}
\item Compute $\theta=\frac{m_1(Y_1)}{m_1(Y_1)+m_2(Y_2)}$\;,
\item Generate a random number $x$ uniformly distributed on $[0,1]$,
\item If $x<\theta$, set $X_n=Y_1$\,; otherwise, set $X_n=Y_2$\,,
\end{enumerate}
\end{enumerate}
\item For any $X\in\genFrame\Theta$\,, compute $\widehat{m}(X)$ by:
$$
\widehat{m}(X)=\frac1{N} \sum_{n=1}^N I[X_n=X]\;.
$$
\end{enumerate}
The following table compares the empirical estimates of $m$, computed by means of particle clouds of different sizes $N$:
$$\begin{array}{@{}l||l|l|l|l|l|l|l|l||l@{}}
\log_{10}N&1&2&3&4&5&6&7&8&\infty
\\\hline
m(\{a\})&0.7&0.41&0.39&0.388&0.382&0.384&0.385&0.385 &0.385
\\
m(\{b\})&0.1&0.08&0.045&0.041&0.040&0.040&0.040&0.040 &0.040
\\
m(\{a,b\})&0&0.01&0.008&0.008&0.008&0.008&0.007&0.007 &0.007
\\
m(\{c\})&0.2&0.13&0.19&0.198&0.200&0.199&0.199&0.199 &0.199
\\
m(\{a,c\})&0&0.12&0.108&0.121&0.121&0.120&0.120&0.120 &0.12
\\
m(\{b,c\})&0&0.25&0.259&0.244&0.249&0.249&0.249&0.249 &0.249
\end{array}$$
\subsection{Comparative tests}
\paragraph{Example 4.}
It is assumed $3$ bbas $m_{1:3}$ on $\genFrame\Theta$ defined by:
$$
m_1(\{a,b\})=m_2(\{a,c\})=m_3(\{c\})=1 \;.
$$
The bbas $m_1$ and $m_3$ are incompatible.
However, $m_2$ is compatible with both $m_1$ and $m_3$\,, which implies that a partial consensus is possible between $m_1$ and $m_2$ or between $m_2$ and $m_3$\,.
As a consequence, PCR$\sharp$ should provide better answers by allowing partial combinations of the bbas. 
The fusion of the $3$ bbas are computed respectively by means of DST, PCR$6$ and PCR$\sharp$\,, and the results confirm the intuition:
{\small\begin{itemize}
\item $z_{\mbox{\tiny DST}}=1$ and $m_{\mbox{\tiny DST}}$ is undefined,
\item $m_{\mbox{\tiny PCR6}}(\{a,b\})=m_{\mbox{\tiny PCR6}}(\{a,c\})=m_{\mbox{\tiny PCR6}}(\{c\})=\frac13$\,,
\item $m_{\mbox{\tiny PCR$\sharp$}}(\{a\})=m_{\mbox{\tiny PCR$\sharp$}}(\{c\})=\frac12$ derived from the consensus $\{a,b\}\cap\{a,c\}\;,$ $\{a,c\}\cap\{c\}$ and their beliefs $m_1(\{a,b\})m_2(\{a,c\})\;,$ $m_2(\{a,c\})m_3(\{c\})$.
\end{itemize}}
\paragraph{Example 5.}
It is assumed $3$ bbas $m_{1:3}$ on $\genFrame\Theta$ defined by:
$$\begin{array}{@{}l@{}}\displaystyle
m_1(\{a\})=0.6\,,\ 
m_1(\{a,b\})=0.4\,,
\vspace{4pt}\\\displaystyle
m_2(\{a\})=0.3\,,\ 
m_2(\{a,c\})=0.7\,,
\vspace{4pt}\\\displaystyle
m_3(\{b\})=0.8\,,\  
m_3(\{a,b,c\})=0.2\,.
\end{array}$$
The computation of PCR$\sharp$ is done step by step:
\\[3pt]\emph{Full consensus.} Full functional consensus are:
{
$$\begin{array}{@{}c@{}||@{}c@{}|@{}c@{}|@{}c@{} |@{}c@{}}
Y_1&\{a,b\}&\{a,b\}&\{a\}& \{a\}
\\\hline
Y_2&\{a,c\}&\{a\}&\{a,c\}& \{a\}
\\\hline
Y_3&\{a,b,c\}&\{a,b,c\}&\{a,b,c\}& \{a,b,c\}
\\\hline
\bigcap_iY_i&\{a\}&\{a\}&\{a\}& \{a\}
\\\hline
\prod_im_i(Y_i)&0.056&0.024&0.084&0.036
\end{array}$$}
\emph{Partial consensus sized $2$.}
Then the possible partial consensus are:
{
$$\begin{array}{@{}c||c|c|c|c@{}}
Y_1&\{a,b\}&\{a,b\}&\{a\}&\{a\}
\\\hline
Y_2&\{a,c\}&\{a\}&\{a,c\}&\{a\}
\\\hline
Y_3&\{b\}&\{b\}&\{b\}&\{b\}
\\\hline
Y_1\cap Y_2&\{a\}&\{a\}&\{a\}&\{a\}
\\\hline
Y_2\cap Y_3&\emptyset&\emptyset&\emptyset&\emptyset
\\\hline
Y_3\cap Y_1&\{b\}&\{b\}&\emptyset&\emptyset
\\\hline
\frac{m_2(Y_2)}{m_2(Y_2)+m_3(Y_3)}&0.467&0.273&1&1
\\\hline
\frac{m_3(Y_3)}{m_2(Y_2)+m_3(Y_3)}&0.533&0.727&0&0
\\\hline
\prod_im_i(Y_i)&0.224&0.096&0.336&0.144
\end{array}$$}
Notice that there is never a $2$-sized consensus involving the pair $(Y_2,Y_3)$\,.
As a consequence, the belief ratios for the partial consensus, \emph{i.e.}:
$$
\omega_\gamma=\frac{\displaystyle
I\left[\bigcap_{i\in\gamma}Y_i\ne\emptyset\right]\prod_{i\in\gamma}m_i(Y_i)
}{\displaystyle
\sum_{\gamma'\in C[2|3]}
I\left[\bigcap_{i\in\gamma'}Y_i\ne\emptyset\right]\prod_{i\in\gamma'}m_i(Y_i)
}
\quad\mbox{for }\gamma\in C[2|3]\;,
$$
 are simplified as follows:
$$\left\{\begin{array}{l@{}}
\omega_{\{1,2\}}=\frac{m_1(Y_1)m_2(Y_2)}{m_1(Y_1)m_2(Y_2)+m_3(Y_3)m_1(Y_1)}=\frac{m_2(Y_2)}{m_2(Y_2)+m_3(Y_3)}\;,
\vspace{5pt}\\
\omega_{\{1,3\}}=\frac{m_3(Y_3)m_1(Y_1)}{m_1(Y_1)m_2(Y_2)+m_3(Y_3)m_1(Y_1)}=\frac{m_3(Y_3)}{m_2(Y_2)+m_3(Y_3)}\;,
\end{array}\right.$$
The case $\gamma=\{2,3\}$ does not hold.
\\[5pt]
\emph{$1$-sized consensus.}
There is no remaining $1$-sized consensus.
\\[5pt]
\emph{Belief compilation.}
The different cases resulted in only two propositions, \emph{i.e.} $\{a\}$ and $\{b\}$.
By combining the entry beliefs $\prod_im_i(Y_i)$ and ratio beliefs, the fused bba $m=m_{\mbox{\tiny PCR$\sharp$}}$ is then deduced:
$$\begin{array}{@{}l@{}}
m(\{a\})= 0.056 + 0.024 + 0.084 + 0.036
+ 0.467\times0.224
\\\rien\qquad\qquad\qquad\qquad\qquad
+ 0.273\times0.096
+ 1\times0.336 + 1\times0.144 =0.811\;,
\\[5pt]
m(\{b\})= 0.533\times0.224 + 0.727\times0.096
=0.189\;.
\end{array}$$
As a conclusion:
$$
m_{\mbox{\tiny PCR$\sharp$}}(\{a\}) = 0.811
\quad\mbox{and}\quad
m_{\mbox{\tiny PCR$\sharp$}}(\{b\}) = 0.189
\;.
$$
It is noticed that $z=0$ for this general case of PCR$\sharp$.
Then, the estimate $\widehat{m}$ is obtained by the following process, working for any choice of $m_{1:3}$\,:
\begin{enumerate}
\item Repeat from $n=1$ to =$n=N$:
\begin{enumerate}
\item Generate $Y_1$\,, $Y_2$ and $Y_3$ by means of $m_1$\,, $m_2$ and $m_3$ respectively,
\item If $Y_1\cap Y_2\cap Y_3\ne\emptyset$\,, then set $X_n=Y_1\cap Y_2\cap Y_3$
and return,
\item If $(Y_1\cap Y_2)\cup(Y_1\cap Y_3)\cup(Y_2\cap Y_3)\ne\emptyset$\,, then do:
\begin{enumerate}
\item For any $\gamma\in C[2|3]=\bigl\{\{1,2\},\{1,3\},\{2,3\}\bigr\}$\,,
do:
\begin{enumerate}
\item If $\displaystyle\bigcap_{i\in\gamma}Y_i=\emptyset$, then set $\omega_gamma=0$\,, 
\item Otherwise, set $\displaystyle\omega_gamma=\prod_{i\in\gamma}m_i(Y_i)$\;,
\end{enumerate}
\item For any $\gamma\in C[2|3]$, set $\omega_gamma=\frac{\omega_gamma}{\sum_{\gamma'\in C[2|3]}\omega_{gamma'}}$\,,
\item Choose $\gamma\in C[2|3]$ randomly accordingly to the probability $\omega$\,,
\item Set $\displaystyle X_n=\bigcap_{i\in\gamma}Y_i$
\item return,
\end{enumerate}
\item Otherwise, do:
\begin{enumerate}
\item Compute $\displaystyle\omega_i=\frac{m_i(Y_i)}{m_1(Y_1)+m_2(Y_2)+m_3(Y_3)}$\;,
\item Choose $k\in\{1,2,3\}$ randomly accordingly to the probability $\omega$\,,
\item Set $X_n=Y_k$\,,
\item return,
\end{enumerate}
\end{enumerate}
\item For any $X\in\genFrame\Theta$\,, compute $\widehat{m}(X)$ by:
$$
\widehat{m}(X)=\frac1{N} \sum_{n=1}^N I[X_n=X]\;.
$$
\end{enumerate}
The following table compares the empirical estimates of $m$, computed by means of particle clouds of different sizes $N$:
$$\begin{array}{@{}l||l|l|l|l|l|l|l|l||l@{}}
\log_{10}N&1&2&3&4&5&6&7&8&\infty
\\\hline
m(\{a\})&1&0.77&0.795&0.812&0.812&0.811&0.811&0.811 &0.811
\\
m(\{b\})&0&0.23&0.205&0.188&0.188&0.189&0.189&0.189 &0.189
\end{array}$$
These results could be compared to DST and PCR6:
{\small\begin{itemize}
\item $z_{\mbox{\tiny DST}}=0.8$ and $m_{\mbox{\tiny DST}}(\{a\})=1$,
\item $m_{\mbox{\tiny PCR6}}(\{a\})=0.391$\,, $m_{\mbox{\tiny PCR6}}(\{b\})=0.341$\,,\\
$m_{\mbox{\tiny PCR6}}(\{a,b\})=0.073$\,, $m_{\mbox{\tiny PCR6}}(\{a,c\})=0.195$\,,
\end{itemize}}
DST produces highly conflicting results, since source $3$ conflicts with the other sources.
However, there are some partial consensus which allow the answer $\{b\}$\,.
DST is blind to these partial consensus.
On the other hand, PCR6 is able to handle hypothesis $\{b\}$\,, but is too much optimistic and, still, is unable to fuse partial consensus.
Consequently, PCR6 is also unable to diagnose the high inconstancy of belief $m_3(\{b\})=0.8$\,.
%
%
\section{Conclusion}
\label{DSmB3:2:F2K9:sect:conclude}
This chapter has investigated a new framework for the definition and interpretation of fusion rule of evidences.
This framework is based on the new concept of referee function.
A referee function models an arbitrament process conditionally to the contributions of several independent sources of information.
It has been shown that fusion rules based on the concept of referee functions have a straightforward sampling-based implementation.
As a consequence, a referee function has a natural algorithmic interpretation.
Owing to the algorithmic nature of referee functions, the conception of new rules of fusion is made easier and intuitive.
Examples of existing fusion rules have been implemented by means of referee functions.
Moreover, an example of rule construction  has been provided on the basis of an arbitrament algorithm.
The new rule is a quite general extension of both PCR6 and Dempster-Shafer rule.
This chapter also addresses the issue of fusion rule approximation.
There are cases for which the fusion computation is prohibitive.
The sampling process implied by the referee function provides a natural method for the approximation and the computation speed-up.
There are still many questions and improvements to be addressed.
For example, samples regularization techniques may reduce possible samples degeneracy thus allowing smaller particles clouds.
Some theoretical questions are also pending; especially, the algebraic properties of the referee functions have almost not been studied.
However, this preliminary work is certainly promising for future applications.
\end{document}